%% file: TrustfulLLMs.tex
\pdfoutput=1

\documentclass[11pt]{article}

\usepackage{acl}

\usepackage{times}
\usepackage{latexsym}

\usepackage[T1]{fontenc}

\usepackage[utf8]{inputenc}

\usepackage{microtype}

\usepackage{inconsolata}

\usepackage{booktabs} 
\usepackage{mathtools}
\usepackage{subcaption}
\usepackage{caption}
\usepackage{amsmath,amssymb,amsfonts}
\usepackage{algorithm} 
\usepackage{algpseudocode}
\usepackage{graphicx}
\usepackage{multirow}

\usepackage{enumitem}
\usepackage{tcolorbox}
\usepackage{mdframed}

\author{Xiaofeng Zhu \\
  Microsoft Corporation / WA, USA\\
  \texttt{Xiaofeng.Zhu@microsoft.com} \\\And
  Jaya Krishna Mandivarapu \\
  Microsoft Corporation / GA, USA\\
  \texttt{jmandivarapu@microsoft.com} \\}

\title{Trustful LLMs: Customizing and Grounding Text Generation with Knowledge Bases and Dual Decoders}

\begin{document}

\maketitle

\begin{abstract}
Although people are impressed by the content generation skills of large language models, the use of LLMs, such as ChatGPT, is limited by the domain grounding of the content. The correctness and groundedness of the generated content need to be based on a verified context, such as results from Retrieval-Augmented Generation (RAG). One important issue when adapting LLMs to a customized domain is that the generated responses are often incomplete, or the additions are not verified and may even be hallucinated. Prior studies on hallucination detection have focused on evaluation metrics, which are not easily adaptable to dynamic domains and can be vulnerable to attacks like jail-breaking. In this work, we propose 1) a post-processing algorithm that leverages knowledge triplets in RAG context to correct hallucinations and 2) a dual-decoder model that fuses RAG context to guide the generation process.
\end{abstract}

\maketitle


\input{TrustfulLLMssections/1.Introduction}

Our contributions are as follows.\\
1) We correct hallucinations and out-of-domain outputs in generated texts from LLMs by leveraging a graph algorithm and provide reasoning using knowledge triplets extracted from both the guided context and the generated texts.\\
2) We propose a dual-decoder model that fuses guided context with natural language generation models, in which the decoders share the weights of a pre-trained LLM.\\
3) The proposed algorithm and model reduce the constraints on the maximum output length, in addition to correcting hallucinations, by returning or generating only outputs related to the prompt and the guided context.\\

\input{TrustfulLLMssections/2.backgroundRelatedWork}

\begin{figure*}[pt]
\begin{center}
\includegraphics[scale=0.55]{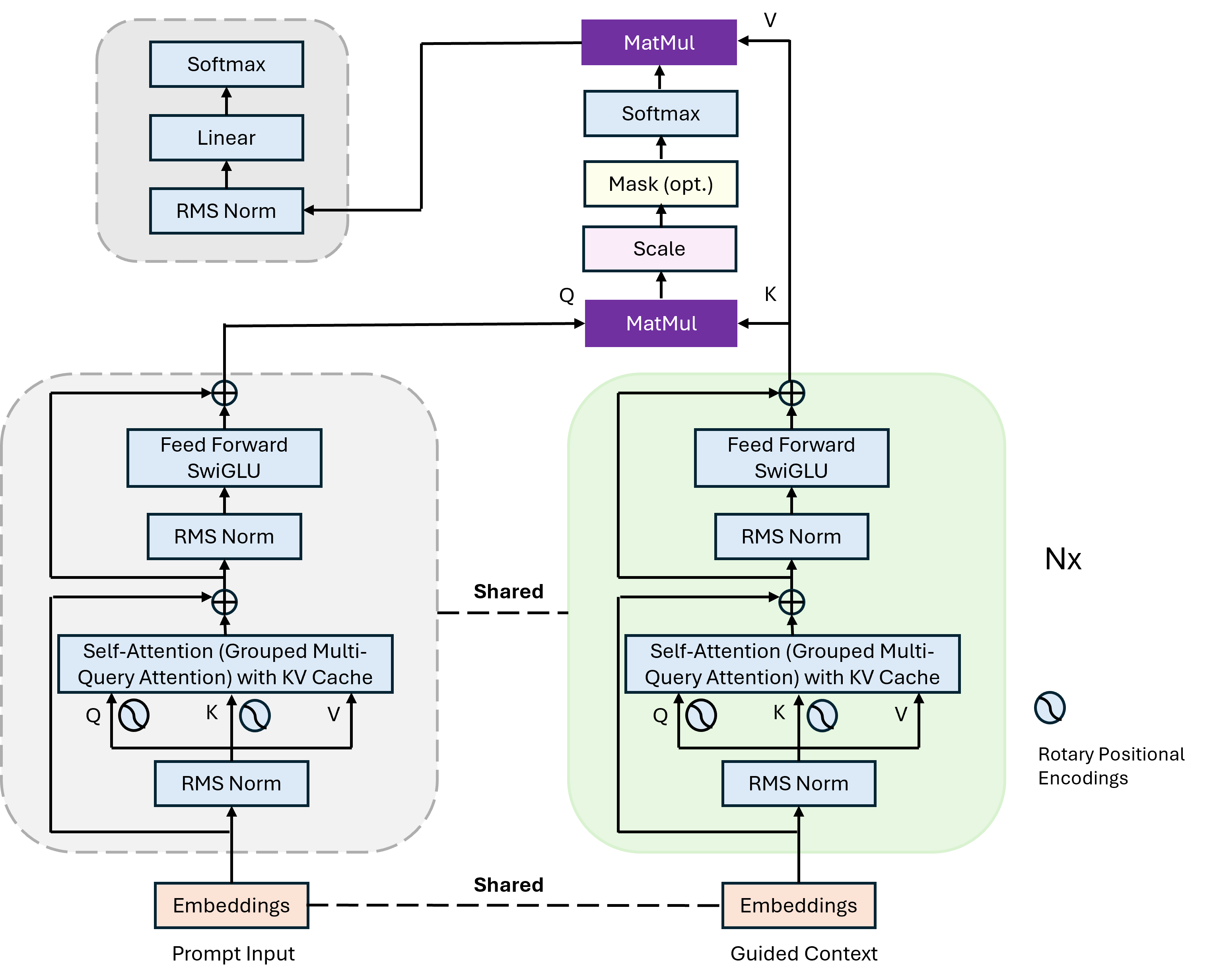}
\caption{TrustfulLLM}
\label{TrustfulLLMs model}
{The dual decoder module can be adapted to any generic LLM, and the weights are shared for the guided context and the prompt input.}
\end{center}
\end{figure*}

\input{TrustfulLLMssections/3.Methodology}


\begin{table*}[t]
\centering
\scalebox{0.8}{
\begin{tabular}{l|r|r|r|r|r}
\multicolumn{6}{c}{M365} \\ \hline\hline
Models 	& ROUGE-L 	& METEOR 	& Groundedness 	& 	GPT-Similarity	& BERTScore   \\ \hline
TrustfulLLM + HC + Phi-3.5-mini-instruct    &\textbf{0.55} &\textbf{0.51} &\textbf{5.00}  &\textbf{4.68}  &\textbf{0.93}   \\ \hline
TrustfulLLM + Phi-3.5-mini-instruct    &0.50 &0.50 &3.98  &4.30  &0.90   \\ \hline
HC + Phi-3.5-mini-instruct    &0.46 &0.48 &\textbf{5.00}  &4.52  &0.91   \\ \hline
RAG + Phi-3.5-mini-instruct  &  0.41& 0.45& 3.72 & 3.49 & 0.89 \\ \hline
RAG + Mistral-NeMo-Minitron-8B-Instruct &   0.38&  0.46  & 3.77 & 3.76&0.88 \\ \hline
RAG + Llama-3.1-8B-Instruct &  0.40& 0.46& 3.74 &3.34 & 0.89 \\ \hline
RAG + GPT-3.5 Turbo &  0.45 & 0.48 &  3.81  & 3.58  &0.90                      \\ \hline
RAG + GPT-4o &  0.42 & 0.48 &  3.77  & 3.52  &0.91                     \\ \hline
Phi-3.5-mini-instruct&  0.17& 0.26& 3.33 &3.60  & 0.84\\ \hline
Mistral-NeMo-Minitron-8B-Instruct&  0.16& 0.24& 3.50 & 4.05 & 0.82 \\ \hline
Llama-3.1-8B-Instruct &  0.19& 0.26&3.44  &3.82  & 0.84 \\ \hline
GPT-3.5 Turbo &  0.23& 0.31& 3.70 & 4.10 & 0.85 \\ \hline
GPT-4o &  0.16& 0.25& 3.64 &3.97  & 0.83 \\ \hline

\end{tabular}
}
\caption{Question Answering Benchmarking Results}
\label{results}
\end{table*}

\input{TrustfulLLMssections/4.ExperimentsResults}

\input{TrustfulLLMssections/5.Conclusion}

\section{Acknowledgments}
We greatly appreciate the help from Dr. Cheng Wu at Microsoft, especially regarding the productization feedback and computational resource support.

\bibliography{ref}

\appendix
\section{Appendix}

\input{TrustfulLLMssections/6.Appendix}

\end{document}

%% file: TrustfulLLMssections/1.Introduction.tex
\section{Introduction}
\label{Introduction}





Adapting an LLM to a specific domain is challenging for several reasons: 1) Pre-trained LLMs cover general knowledge and cannot access private data (even during fine-tuning) due to privacy, copyright, and policy constraints. 2) The grounding of generated texts can change depending on specific contexts, such as domain or timestamp. Recent studies mostly focus on detecting hallucinations and using multiple LLMs when hallucinations occur. 3) Business logic and structured data, such as databases and private knowledge bases, are required when integrating customized LLMs into production systems and presenting them to customers or users.

We offer two methods for correcting hallucinations (beyond merely detecting them \cite{wan2024acueval, li2023trustworthy, 10.1145/3571730}): 1) Applying post-processing to generated texts using knowledge triplets, and 2) Proposing guided generation via Dual Decoders. Inspired by common practices like Retrieval-Augmented Generation (RAG) \cite{li2024retrieval}, which retrieves relevant grounding context and feeds it to an LLM for text generation, we address hallucinations in generated texts from two aspects: 1) Post-editing based on knowledge graphs extracted from the context, and 2) Infusing guided context that contains important knowledge triplets into a generic LLM. Our proposed methods also provide reasoning and create consistent results from generative LLMs, benefiting from both the generation and extraction capabilities of decoder-only LLMs and the groundedness of RAG via the second decoder on the guidance \cite{le2020dualdecodertransformerjointautomatic, Wang_2022}.

In this work, we elaborate on our real-world commercial application scenario of using LLMs to support customers with Microsoft product inquiries in copilots, where groundedness is key to success. Pre-trained LLMs often lack the relevant knowledge or cannot adapt promptly to changes in the product database updates.
Different variants of large language models (LLMs), such as Phi-3.5 \cite{abdin2024phi3technicalreporthighly}, ChatGPT \cite{mohamadi2023chatgptagegenerativeai}, LLama-3 \cite{dubey2024llama3herdmodels}, and Gemma \cite{gemma_2024}, are proficient at producing fluent outputs for diverse user queries. Despite their human-like fluency in generating text across a wide range of prompts, large language models suffer from \textit{hallucinations} (see examples in Figures \ref{Microsoft-Business-Basic}, \ref{GPT-3.5-Turbo}, \ref{Phi-3}), where parts or the entirety of the generated text lack faithfulness, factuality, or reasoning, yet are presented with a confident tone \citealp{10.1145/3571730}.\\

To mitigate and correct hallucinations, we leverage guided text generation. Grounding guidance \cite{socher2013reasoning, nickel2011three, lin2015learning, wang2014knowledge, bordes2013translating, wang2022language, grover2016node2vec}, such as knowledge graphs (KGs), has been shown to significantly improve the reliability and factuality of LLMs in recent studies, e.g., KELM \cite{agarwal2020knowledge, lu2021kelm}, SKILL \cite{moiseev2022skill}, K-DLM \cite{zou2023k}, KEPLET \cite{li2023keplet}, and LUKE-Graph \cite{foolad2023luke}. Knowledge graphs typically consist of factual information represented explicitly in a semi-structured format, generally as [subject entity, relation, object entity] triples, e.g., (Bill Gates, was, the CEO of Microsoft) \cite{han-etal-2019-opennre, Gardner2017AllenNLP}. We collect and maintain such knowledge triplets and grounded context offline for RAG.

%% file: TrustfulLLMssections/2.backgroundRelatedWork.tex
\section{Background and Related Work}
\label{Background and Related Work}

Unlike document summarization, RAG, or traditional question answering, our approach benefits from both domain knowledge bases—particularly for groundedness—and the language understanding and generalization capabilities of various pre-trained or customized LLMs. By iterating over the knowledge triplets extracted from the generated text and comparing them to the knowledge triplets extracted from the given context (e.g., results from RAG), we can correct hallucinations (and generated phrases that lack references) using our proposed post-processing algorithm.

\subsection{Guided Natural Language Generation}
Prior studies have attempted multiple guidance frameworks, particularly with encoder-decoder models \cite{see2017get, dou2020gsum, hokamp2017lexicallyconstraineddecodingsequence, beurerkellner2024guidingllmsrightway,10.1145/3543873.3584621,LINEARDIFFERENCEVECTOR}. Unlike GraphRAG \cite{edge2024localglobalgraphrag}, which utilizes multiple LLM calls to combine knowledge triplets from segments of RAG results, our proposed TrustfulLLM model reduces irrelevant entities and tokens in generated texts to demonstrate its efficiency.\\

\subsection{Hallucination}
Hallucination is considered one of the most prominent drawbacks of Large Language Models, as it leads models to generate inaccurate or false information \cite{10.1145/3571730, wan2024acueval}. Model-generated texts may not match the true source content, and the facts presented by the model cannot always be verified from the source. These drawbacks remain significant hurdles in applying large language models (LLMs) to real-world, business-critical, and vitally important applications.

%% file: TrustfulLLMssections/3.Methodology.tex
\begin{algorithm}
	\caption{Hallucination Correction}
	\begin{algorithmic}[1]
            \State \textbf{Input}: $\hat{Y}$, $G$
            \State \textbf{Output}: $Y^*$
            \State Construct knowledge graph $g = \{t_i\}$ from $\hat{Y}$
		\For {knowledge triplet $t_i = (v^s_i, v^o_i, r_i)$ in $g$}
                \If {$v^s_i$ not in $G$}
                    \State Eliminate $t_i$ from $g$ and the associated sentence in $\hat{Y}$
                \Else
                    \State Replace/keep $t_i$ in $\hat{Y}$ based on $g$ and $G$
                \EndIf
		\EndFor
            \State Assume $\hat{G}$ is the subgraph of $G$, and $\hat{G}$ contains all the verified entities (nodes) in $\hat{Y}$
            \State $Y^*$ = $\hat{Y}$
            \While {$Y^*$ contains cycles}
            \State Prune $\hat{Y}$ to $Y^*$ till $Y^*$ is a minimum spanning tree of $\hat{G}$.
            \EndWhile
	\end{algorithmic}
        \label{Algorithm HC}
\end{algorithm}

\section{Methodology}
\label{Methodology}
Whether the generated text is factual is determined by the domain source and the given guided context. In our copilot scenario, we always retrieve related context for a user prompt/query and then utilize this context to generate the final response presented to users. The guided context can be a mix of offline or web articles and database records, from which we generate knowledge triplets \cite{Gardner2017AllenNLP} for groundedness verification and hallucination correction. We propose a post-processing algorithm for correcting hallucinations that can be applied to any LLM outputs, as discussed in Section \ref{HC}. Additionally, we propose a dual-decoder text generation model that takes both the prompt and guided context leveraging the RAG result content as inputs, described in Section \ref{TrustfulLLMs}. 

\subsection{Post-processing text generation by Correcting Knowledge Triplets}
\label{HC}
For generated texts from an LLM, we identify and correct potential hallucinations using knowledge triplets extracted from the RAG context and the generated text output. Specifically, we convert the extracted knowledge triplets from the guided context and the LLM output into graphs $G$ and $g$, respectively, where each node $v_i$ represents either a subject or an object, and the relations between the subject and object serve as bi-directional edges connecting the two nodes\cite{8102916}. Algorithm \ref{Algorithm HC} explains the hallucination detection and correction process for a given generated text $\hat{Y}$ and the knowledge graph $G$ extracted from the guided context. In the end, we obtain a corrected/verified output $Y^*$. A knowledge triplet $t$ can be identified given a subject and a relation, or an object and a relation; i.e., we can easily locate and replace the third component when the entity or relation is incorrect in $t_i$, which is composed of subject $v^s_i$, object $v^o_i$, and the relation $r_i$. This algorithm can verify, replace, and prune triplets in $\hat{Y}$ but does not increase the number of nodes/entities. For instance, given a sentence in RAG result content: \textit{"Microsoft 365 Business Basic is \$7.2 dollars per user per month."}, we obtain knowledge triplet $t_i$: ($v^s_i, v^o_i, r_i$) is \textit{(Microsoft 365 Business Basic, is, \$7.2 dollars per user per month)}. Since LLM outputs can omit/add entities, we propose a second method: guided generation via dual decoders, which fundamentally alters the text generation process.

\subsection{TrustfulLLM and Guided Generation via Dual Decoders}
\label{TrustfulLLMs}
In addition to the contextual embeddings used in Transformers, we embed the guidance text and apply a cross-attention calculation using the hidden states of the two decoders. In this way, we have the grounding/context source embeddings in one decoder and the user prompt in the other decoder, with both decoders sharing weights. We apply cross-attention CROSSATTN$(H_p, H_g)$ by taking the hidden state $H_p$ of the prompt module as the `query' and the hidden state $H_g$ of the guided context module as the `key' and `value.' The diagram of the TrustfulLLM is shown in Figure \ref{TrustfulLLMs model}, and the pre-trained LLM component is generic. Only the prompt inputs are generated token by token, while the guided context contributes to the CROSSATTN$(H_p, H_g)$ only. The fine-tuned transformer block components (the grey boxes in Figure \ref{TrustfulLLMs model}) are derived from the Phi-3 and model architecture \cite{abdin2024phi3technicalreporthighly, dubey2024llama3herdmodels, vaswani2023attentionneed}.

During the inference stage, the guided context is the same as the RAG context. We augment the RAG context by randomly adding additional content (shuffled from other RAG results from different prompts) as the guided context during fine-tuning, as shown in the Appendix \ref{Examples of Prompt, RAG Context, and Guided Context}.


%% file: TrustfulLLMssections/4.ExperimentsResults.tex
\section{Experiments and Results}
\label{Experiments and Results}
\subsection{\textbf{Tasks and Datasets}}
We elaborate the results from the public Microsoft learn.microsoft.com articles and product from www.microsoft.com \footnote{\url{https://github.com/MicrosoftDocs/microsoft-365-docs}}.
The M365 dataset comprises approximately 10,000 question-and-answer pairs, including the context from which these question and answers were derived. We conducted our experiments based on that the RAG results (knowlege bases and/or domain articles) that are trustworthy. For fine-tuned LLMs, we leverage LoRA \cite{hu2021loralowrankadaptationlarge} and set the number of epochs to be over 400, which is relatively higher than in regular LoRA fine-tuning.

\subsection{Metrics and Baseline Models}

We use a combination of metrics including ROUGE-L, METEOR, GPT-Similarity, Groundedness (Appendix \ref{sec:GROUNDNESS_SIMILARITY_PROMPTS}), and BERTScore. ROUGE-L assesses the longest common subsequence between the generated and reference texts, capturing fluency and coherence. METEOR goes further by considering synonyms, stemming, and word order, providing a more nuanced evaluation. Groundedness rated 1-5 by GPT-4 ensures that the generated content is closely aligned with the source material. GPT-Similarity rated 1-5 by GPT-4 measures the semantic similarity between generated and reference texts, while BERT Score leverages pre-trained language models to evaluate the quality of the generated text on a deeper, contextual level. Together, these metrics provide a comprehensive assessment of our model performance. 

We show the results of our methods, pre-trained LLMs, RAG, and Trustful LLMs on domain datasets M365 in Table \ref{results}, where boldface indicates the best scores, HC indicates applying the hallucination correction post-processsing algorithm, and TrustfulLLM indicates fine-tuning from the pre-trained model on the domain data. Leveraging the proposed HC can largely boost the groundedness score, and utilizing the TrustfulLLM dual-decoder framework and HC yield the best performance among all metrics. 
In particular, the percentage of eliminated entities when HC is applied to Phi-3.5 decreases from 18\% to 6.9\% when HC is applied to TrustfulLLM + Phi-3.5, further supporting the effectiveness of TrustfulLLM. We also explored the performance of the models on a general summarization task in Appendix \ref{Summarization Task}.

\subsection{Effects of Applying HC and TrustfulLLM}
\label{Effects of Applying HC and TrustfulLLM}
We take a incorrect \& incomplete statement from an LLM as a straightforward example: \textit{"Domain registrar that support all DNS records required for Microsoft 365 are GoDaddy and Oray."} After we apply HC, HC corrects this output as follows: \textit{"Domain registrars that support all DNS records required for Microsoft 365 are Oray , HiChina , east.net, and BIZCN."}

In our production systems, we convert the nodes at Line 4 of Algorithm \ref{Algorithm HC} into embeddings using a pre-trained transformer model, allowing us to find semantically related subjects/objects using the cosine similarity and a heuristic similarity threshold. For example, \textit{"M365 Business Basic"} can be mapped to \textit{"Microsoft 365 Business Basic"}. When offline \& pre-calibrated knowledge triplets are available, especially for user prompts related to Microsoft product information, we store the embeddings using the FAISS\cite{douze2024faiss} \footnote{https://github.com/facebookresearch/faiss} and combine them with the knowledge triplets extracted in the real-time RAG context.

LLMs can generate content that does not originate from the RAG context, which may not always be a hallucination. However, HC can make the outputs more consistent and better aligned with the RAG \& guided context. For instance, given a user prompt: 

\textit{How much is M365 Business Basic?}

and RAG context: 

\textit{"Microsoft 365 Business Basic is \$7.2 dollars per user per month. If you commit yearly the price is \$6 dollars per user per month. Microsoft 365 Business Standard is \$15 dollars per user per month ..."},

a LLM can respond with: 

\textit{Microsoft 365 Business Basic is priced at \$7.2 dollars for each user on a monthly basis. However, if you choose to commit to a yearly plan, the price decreases to \$6 dollars per user per month.} 

after adding the below RAG context to the prompt:

\textit{Please answer using the exact following context "Microsoft 365 Business Basic is \$7.2 dollars per user per month. If you commit yearly the price is \$6 dollars per user per month. Microsoft 365 Business Standard is \$15 dollars per user per month ..."}.

Utilizing TrustfulLLM and HC, we can obtain the following outputs: 

\textit{"Microsoft 365 Business Basic is \$7.2 dollars per user per month. If you commit yearly the price is \$6 dollars per user per month."}

Apparently, TrustfulLLM and HC tends to use the knowledge triplets and exact sub-sentences from the verified RAG context for grounded generation. This also makes (human) evaluations more effective in production systems.

\subsection{Commercial Application and Constraints}
In our commercial system, we first apply an intent detection to user prompts to filter out enquiries that are not related to our business then apply a retrieval model to obtain most relevant internal documents, records in product databases. We only reply on the groundedness and correctness of the retrieval results, i.e, phrases in AI generated texts that cannot be referenced from the RAG results are eliminated. For phrases that are semantically equivalent to the RAG results we still do a replacement using the knowledge triplet correction to keep consistent responses. We have also thoroughly conducted Red Teaming evaluations on various Responsible AI metrics such as harmful content, IP infringement, jailbreaking, groundedness, etc. Though we highligh our proposed halluciation correction algorithm and the dual decoder architecture, the upstream RAG and intent detection models can be combined in a multi-task modeling process.

%% file: TrustfulLLMssections/5.Conclusion.tex
\section{Conclusion}
\label{Conclusion}
We have addressed grounding issues in LLMs and proposed task-agnostic hallucination correction methods for real-world applications from two perspectives: post-processing to refine LLM outputs and trustful LLM fine-tuning via dual encoders. We have discussed hallucination correction and trustworthy text generation, demonstrating the robustness and resilience of our methods. In the future, we plan to explore heterogeneous modalities, such as structured and spatio-temporal data, knowledge-enriched representations of input tokens \cite{grover2016node2vec, yu-etal-2022-empirical, pan2023unifying, gao2021triples, ye2021contrastive}, hierarchical relation graphs, and accountability \cite{li2023trustworthy}. We also plan to study model bias, aggregation for federated learning \cite{zheng2023fedpse, hashemi2021byzantine}, and privacy-preserving issues \cite{hashemi2021byzantine}. Additionally, we aim to reduce the complexity of LLMs through parameter-efficient fine-tuning.

%% file: TrustfulLLMssections/6.Appendix.tex
\subsection{Hallucination Examples}
We show examples where various LLMs generate hallucinations.

\begin{figure}[ht!]
\includegraphics[scale=0.4]{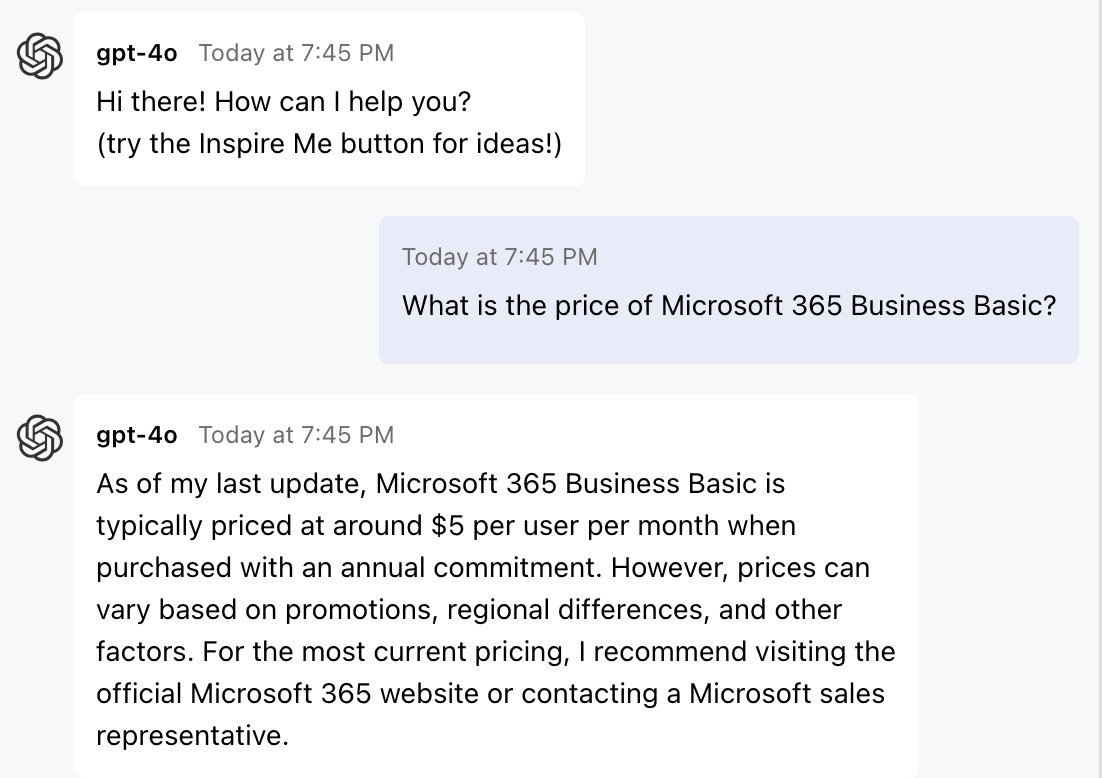}
\caption{Hallucination Example 1}
\label{Microsoft-Business-Basic}
{GPT-4o returns the wrong pricing information of Microsoft 365 Business Basic.}
\end{figure}

\begin{figure}[ht!]
\includegraphics[scale=0.4]{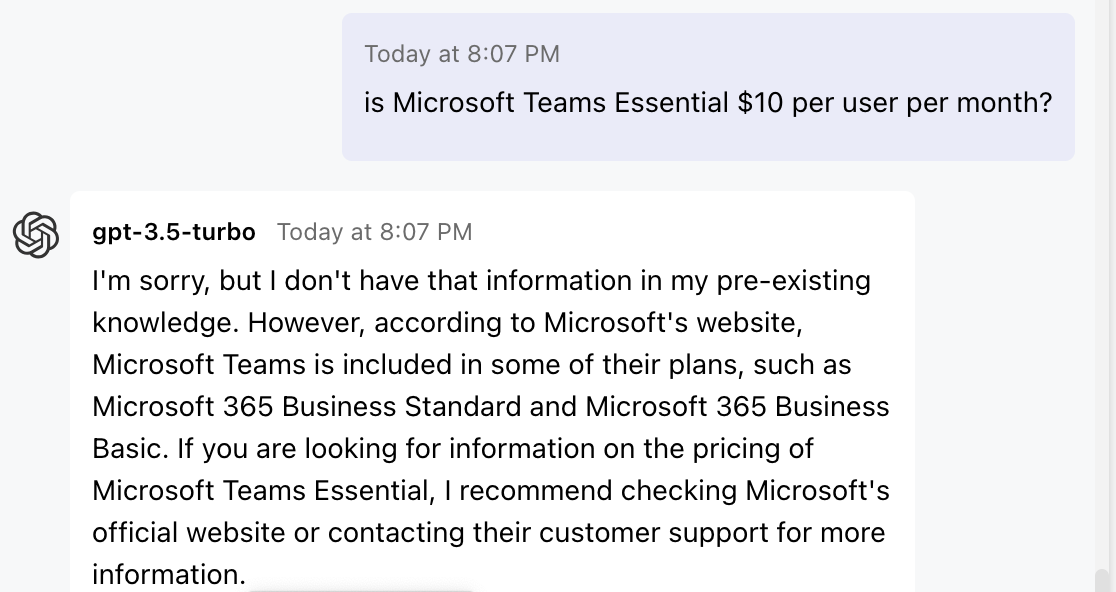}
\caption{Hallucination Example 2}
\label{GPT-3.5-Turbo}
{GPT-3.5 Turbo cannot answer questions related to Microsoft Teams Essential.}
\end{figure}

\begin{figure}[ht!]
\includegraphics[scale=0.4]{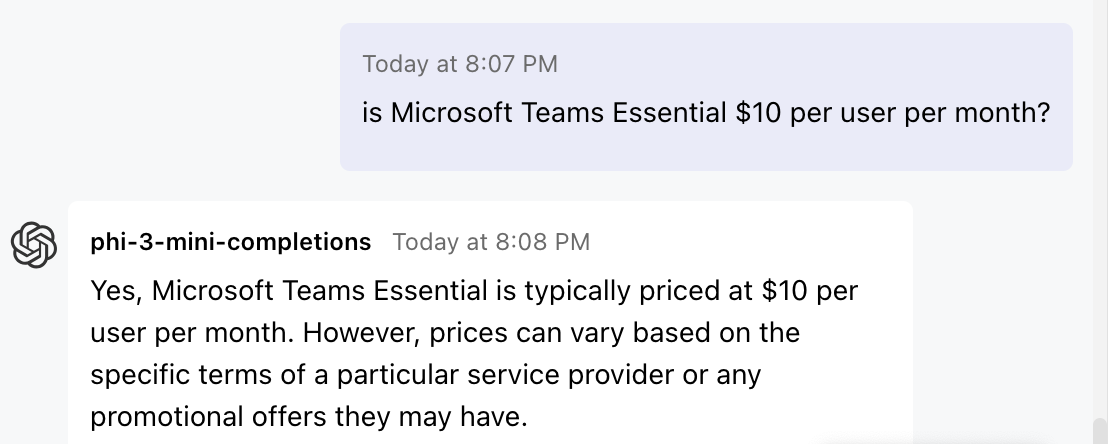}
\caption{Hallucination Example 3}
\label{Phi-3}
{Phi-3 answered incorrectly about the price of Microsoft Teams Essential.}
\end{figure}

\subsection{Examples of Prompt, RAG Context, and Guided Context}
\label{Examples of Prompt, RAG Context, and Guided Context}
Prompt: \textit{"... <|user|> How much is Microsoft 365 Business Basic? <|end|> <|assistant|>  Microsoft 365 Business Basic is \$7.2 dollars per user per month. <|end|>"}.

RAG context: \textit{"Microsoft 365 Business Basic is \$7.2 dollars per user per month. Microsoft 365 Business Basic ..."}.

Guided context: \textit{"Microsoft 365 Business Basic is \$7.2 dollars per user per month. Microsoft 365 Business Basic ... Microsoft 365 Business Standard is ... <|end|>"}. We add additional content about, such as \textit{"Microsoft 365 Business Standard"}, which is similar to the product \textit{"Microsoft 365 Business Basic"} to the RAG context. This is for mimicking the potentially noisy RAG context in the retrieval stage.



\subsection{Summarization Task}
\label{Summarization Task}

A summarization task does not have the retrieval component as in RAG. We utilize the graph building step of HC to select the salient sentences from the articles as the guided context. We first extract knowledge triplets from the articles then keep sentences where the most frequent subjects are associated with. We show the comparison of TrustfulLLM + HC + Phi-3.5-mini-instruct, where HC extract knowledge triplets from the articles and the generated texts in the inference stage, and LLM baselines in Table \ref{results_cnndailymail}.

\begin{table*}[t]
\centering
\scalebox{0.8}{
\begin{tabular}{l|r|r|r|r|r}
\multicolumn{6}{c}{CNN DailyMail} \\ \hline\hline
Models 	& ROUGE-L 	& METEOR 	& Groundedness 	& 	GPT-Similarity	& BERTScore   \\ \hline
TrustfulLLM + HC + Phi-3.5-mini-instruct    &\textbf{0.41} &\textbf{0.39} &\textbf{5.00}  &\textbf{4.12}  &\textbf{0.89}   \\ \hline
TrustfulLLM + Phi-3.5-mini-instruct    &  0.40 & \textbf{0.39} &4.68  &\textbf{4.12}  & 0.88 \\ \hline
HC + Phi-3.5-mini-instruct    &  0.35 & 0.36 &\textbf{5.00} &3.82  & 0.88 \\ \hline
Phi-3.5-mini-instruct &  0.17& 0.34& 4.29 &3.79  & 0.86\\ \hline
Mistral-NeMo-Minitron-8B-Instruct &  0.20& 0.35& 3.32 & 3.87 & 0.86 \\ \hline
Llama-3.1-8B-Instruct &  0.32 & 0.37 &4.61  &4.10  & 0.87 \\ \hline
GPT-3.5 Turbo &  0.24& 0.38& 4.50 & 3.79 & 0.87 \\ \hline
GPT-4o &  0.18& 0.36& 4.42 & 4.10  & 0.87\\ \hline
\end{tabular}
}
\caption{Summarization Benchmarking Results}
\label{results_cnndailymail}
\end{table*}

\subsection{Prompt Template for \textbf{GPT Metrics}}
\label{sec:GROUNDNESS_SIMILARITY_PROMPTS}
We show the prompts of GPT-Similarity and Groundedness addressed in Section \ref{Experiments and Results}.

\begin{mdframed}[linecolor=black]
\textbf{Prompt for  Groundedness}\\
\textbf{\textcolor{red}{System:}}\\ You are an AI assistant. You will be given the definition of an evaluation metric for assessing the quality of an answer in a question-answering task. Your job is to compute an accurate evaluation score using the provided evaluation metric. You should return a single integer value between 1 to 5 representing the evaluation metric. You will include no other text or information.

\noindent\textbf{\textcolor{red}{User:}}\\ You will be presented with a CONTEXT and an ANSWER about that CONTEXT. You need to decide whether the ANSWER is entailed by the CONTEXT by choosing one of the following rating:
\begin{enumerate}
    \item 5: The ANSWER follows logically from the information contained in the CONTEXT.
    \item 1: The ANSWER is logically false from the information contained in the CONTEXT.
    \item An integer score between 1 and 5, and if such an integer score does not exist, use 1: It is not possible to determine whether the ANSWER is true or false without further information.
\end{enumerate}

Read the passage of information thoroughly and select the correct answer from the three answer labels. Read the CONTEXT thoroughly to ensure you know what the CONTEXT entails. Note that the ANSWER is generated by a computer system, so it can contain certain symbols, which should not be a negative factor in the evaluation.

\noindent\textbf{Independent Examples:}\\
\noindent\textbf{Example Task \#1 \textcolor{blue}{Input:}}\\
\{"CONTEXT": "Some are reported as not having been wanted at all.", "QUESTION": "", "ANSWER": "All are reported as being completely and fully wanted."\}\\
\textbf{Example Task \#1 Output:} \\1

\noindent\textbf{Example Task \#2 \textcolor{blue}{Input:}}\\
\{"CONTEXT": "Ten new television shows appeared during the month of September. Five of the shows were sitcoms, three were hourlong dramas, and two were news-magazine shows. By January, only seven of these new shows were still on the air. Five of the shows that remained were sitcoms.", "QUESTION": "", "ANSWER": "At least one of the shows that were cancelled was an hourlong drama."\}\\
\textbf{Example Task \#2 Output:}\\ 5

\noindent\textbf{Example Task \#3 \textcolor{blue}{Input:}}\\
\{"CONTEXT": "In Quebec, an allophone is a resident, usually an immigrant, whose mother tongue or home language is neither French nor English.", "QUESTION": "", "ANSWER": "In Quebec, an allophone is a resident, usually an immigrant, whose mother tongue or home language is not French."\}\\
\textbf{Example Task \#3 Output:} \\5

\noindent\textbf{Example Task \#4 \textcolor{blue}{Input:}}\\
\{"CONTEXT": "Some are reported as not having been wanted at all.", "QUESTION": "", "ANSWER": "All are reported as being completely and fully wanted."\}\\
\textbf{Example Task \#4 Output:} \\1

\textbf{Actual Task \textcolor{blue}{Input:}}\\
\{"CONTEXT": \{\{context\}\}, "QUESTION": "", "ANSWER": \{\{response\}\}\}\\
Reminder: The return values for each task should be correctly formatted as an integer between 1 and 5. Do not repeat the context and question.

\textbf{Actual Task Output:}

\end{mdframed}

\begin{mdframed}[linecolor=black]
\textbf{Prompt for  GPT-Similarity]}\\
\textbf{\textcolor{red}{System:}}\\
You are an AI assistant. You will be given the definition of an evaluation metric for assessing the quality of an answer in a question-answering task. 
Your job is to compute an accurate evaluation score using the provided evaluation metric. 
You should return a single integer value between 1 to 5 representing the evaluation metric. 
You will include no other text or information.

\noindent\textbf{\textcolor{red}{User:}}\\
Equivalence, as a metric, measures the similarity between the predicted answer and the correct answer. If the information and content in the predicted answer 
is similar or equivalent to the correct answer, then the value of the Equivalence metric should be high, else it should be low. Given the question, correct answer, and predicted answer, determine the value of the Equivalence metric using the following rating scale:

\begin{itemize}
    \item One star: the predicted answer is not at all similar to the correct answer
    \item Two stars: the predicted answer is mostly not similar to the correct answer
    \item Three stars: the predicted answer is somewhat similar to the correct answer
    \item Four stars: the predicted answer is mostly similar to the correct answer
    \item Five stars: the predicted answer is completely similar to the correct answer
\end{itemize}

This rating value should always be an integer between 1 and 5. So the rating produced should be 1, 2, 3, 4, or 5. The examples below show the Equivalence score for a question, a correct answer, and a predicted answer.

\textbf{Question:} What is the role of ribosomes?\\
\textbf{Correct answer:} Ribosomes are cellular structures responsible for protein synthesis. They interpret the genetic information carried by messenger RNA (mRNA) and use it to assemble amino acids into proteins.\\
\textbf{Predicted answer:} Ribosomes participate in carbohydrate breakdown by removing nutrients from complex sugar molecules.\\
\textbf{Stars:} 1\\

\textbf{Question:} Why did the Titanic sink?\\
\textbf{Correct answer:} The Titanic sank after it struck an iceberg during its maiden voyage in 1912. The impact caused the ship's hull to breach, allowing water to flood into the vessel. The ship's design, lifeboat shortage, and lack of timely rescue efforts contributed to the tragic loss of life.\\
\textbf{Predicted answer:} The sinking of the Titanic was a result of a large iceberg collision. This caused the ship to take on water and eventually sink, leading to the death of many passengers due to a shortage of lifeboats and insufficient rescue attempts.\\
\textbf{Stars:} 2\\



\textbf{Question:} What are the health benefits of regular exercise?\\
\textbf{Correct answer:} Regular exercise can help maintain a healthy weight, increase muscle and bone strength, and reduce the risk of chronic diseases. It also promotes mental well-being by reducing stress and improving overall mood.\\
\textbf{Predicted answer:} Routine physical activity can contribute to maintaining ideal body weight, enhancing muscle and bone strength, and preventing chronic illnesses. In addition, it supports mental health by alleviating stress and augmenting general mood.\\
\textbf{Stars:} 5\\

\noindent\textbf{Question:} \textcolor{red}{\{\{query\}\}}\\
\textbf{Correct answer:} \textcolor{red}{\{\{ground\_truth\}\}}\\
\textbf{Predicted answer:} \textcolor{red}{\{\{response\}\}}\\
\textbf{Stars:}
\end{mdframed}